%% file: main.tex
\documentclass[10pt,conference]{IEEEtran}
\IEEEoverridecommandlockouts

\newif\ifieee
\ieeefalse

\usepackage{multirow}
\usepackage{booktabs}
\usepackage{float}
\usepackage{amsmath,amssymb,amsfonts}
\usepackage{algorithmic}
\usepackage{graphicx}
\usepackage{textcomp}
\usepackage[dvipsnames,table,xcdraw]{xcolor}
\def\BibTeX{{\rm B\kern-.05em{\sc i\kern-.025em b}\kern-.08em
    T\kern-.1667em\lower.7ex\hbox{E}\kern-.125emX}}
\usepackage{soul} %
\setuldepth{Berlin}
\usepackage{xspace}

\usepackage[caption=false,farskip=0pt]{subfig}
\usepackage[hyphens]{url}
\ifieee
\else
\usepackage[colorlinks=true,allcolors=black]{hyperref}
\fi
\usepackage[acronym]{glossaries}
\usepackage[sort,compress]{cite} %
\usepackage{balance}
\usepackage{arydshln}
\usepackage[T1]{fontenc}
\usepackage[capitalise]{cleveref} %

\newcommand{\indentresults}{\hspace{3mm}}

\newacronymstyle{long-short-br}
{%
  \GlsUseAcrEntryDispStyle{long-short}%
}%
{%
  \GlsUseAcrStyleDefs{long-short}%
}
\setacronymstyle{long-short-br}
    
\newif\iffinal
\finaltrue
\newcommand{\cmtid}{73}
\newcommand{\papertitle}{Combining Attention Module and Pixel Shuffle\\for License Plate Super-Resolution}

\iffinal
\else
\usepackage[switch]{lineno}
\fi

\begin{document}

\iffinal

\title{\papertitle}

\author{Valfride Nascimento\IEEEauthorrefmark{1}, Rayson Laroca\IEEEauthorrefmark{1}, Jorge de A. Lambert\IEEEauthorrefmark{2}, William Robson Schwartz\IEEEauthorrefmark{3}, and David Menotti\IEEEauthorrefmark{1}\\
\IEEEauthorrefmark{1}\hspace{0.2mm}Department of Informatics, Federal University of Paran\'a, Curitiba, Brazil\\
\IEEEauthorrefmark{2}\hspace{0.2mm}Regional Superintendence at Bahia, Brazilian Federal Police, Salvador, Brazil\\
\IEEEauthorrefmark{3}\hspace{0.2mm}Department of Computer Science, Federal University of Minas Gerais, Belo Horizonte, Brazil\\
\resizebox{0.95\linewidth}{!}{
\IEEEauthorrefmark{1}{\tt\small \{vwnascimento,rblsantos,menotti\}@inf.ufpr.br} \quad \IEEEauthorrefmark{2}{\tt\small lambert.jal@pf.gov.br} \quad \IEEEauthorrefmark{3}\hspace{0.3mm}{\tt\small william@dcc.ufmg.br}
}}

\else
    \title{\papertitle}

    \author{SIBGRAPI paper ID: \cmtid }
    \linenumbers
\fi

\newcommand*{\DM}[2][]{\textcolor{red}{[\textbf{\ifthenelse{\equal{#1}{}}{DM}{DM(#1)}}: #2]}}
\newcommand*{\VN}[2][]{\textcolor{blue}{[\textbf{\ifthenelse{\equal{#1}{}}{VN}{VN(#1)}}: #2]}}
\newcommand*{\RL}[2][]{\textcolor{Rhodamine}{[\textbf{\ifthenelse{\equal{#1}{}}{RL}{RL(#1)}}: #2]}}

\maketitle
\ifieee
{\let\thefootnote\relax\footnote{\\978-1-6654-5385-1/22/\$31.00
\textcopyright2022 IEEE}}
\else
\fi

\newcommand*{\todo}[2][]{\textcolor{red}{[\textbf{\ifthenelse{\equal{#1}{}}{TODO}{TODO(#1)}}: #2]}}

\newcommand\red[1]{{\textcolor{red}{#1}}}
\newcommand\orange[1]{{\textcolor{BurntOrange}{\textbf{#1}}}}

\input{0-aux/acronyms}
\input{0-aux/variables}
\input{0-aux/abstract}

\input{1-Introduction/Introduction}

\input{2-RelatedWork/RelatedWork}

\input{3-Methodology/Methodology}

\input{4-Experiments/Experiments}

\input{5-Conclusions/Conclusions}

\input{0-aux/acknowledgments}

\footnotesize
\bibliographystyle{IEEEtran}
\bibliography{bibliography}

\end{document}

%% file: 0-aux/acronyms.tex
\newacronym{cnn}{CNN}{Convolutional Neural Network}
\newacronym{hr}{HR}{high-resolution}
\newacronym{lp}{LP}{license plate}
\newacronym{lpr}{LPR}{License Plate Recognition}
\newacronym{lr}{LR}{low-resolution}
\newacronym{mse}{MSE}{Mean Squared Error}
\newacronym{ocr}{OCR}{Optical Character Recognition}
\newacronym{psnr}{PSNR}{Peak Signal-to-Noise Ratio}
\newacronym{sisr}{SISR}{Single-Image Super-Resolution}
\newacronym{sr}{SR}{super-resolution}
\newacronym{srcnn}{SRCNN}{Super-Resolution Convolutional Neural Network}
\newacronym{ssim}{SSIM}{Structural Similarity Index Measure}
\newacronym{rcb}{RCB}{Residual Concatenation Block}
\newacronym{fm}{FM}{Feature Module}
\newacronym{sfe}{SFE}{Shallow Feature Extractor}
\newacronym{ps}{PS}{\textit{PixelShuffle}}
\newacronym{pu}{PU}{\textit{PixelUnshuffle}}
\newacronym{nn}{NN}{Neural Network}
\newacronym{psfe}{PSFE}{Pre-shallow Feature Extractor}
\newacronym{pltfam}{PLTFAM}{Pixelshuffle Two-fold Attention Module}
\newacronym{ca}{CA}{Channel Unit}
\newacronym{pos}{POS}{Positional Unit}
\newacronym{tfam}{TFAM}{Two-fold Attention Module}
\newacronym{map}{MAP}{Maximum a Posteriori}
\newacronym{gan}{GAN}{Generative Adversarial Network}
\newacronym{ccpd}{CCPD}{Chinese City Parking Dataset}
\newacronym{mprnet}{MPRNet}{Multi-Path Residual Network}
\newacronym{cbam}{CBAM}{Convolution Block Attention Module}
\newacronym{se}{SE}{Squeeze-and-excitation}
\newacronym{esa}{ESA}{Enhanced Spatial Attention}
\newacronym{csrgan}{CSRGAN}{Character-Based Super-Resolution Generative Adversarial Networks}
\newacronym{srgan}{SRGAN}{Super-Resolution Generative Adversarial Networks}

\newcommand{\rodosolalpr}{RodoSol-ALPR\xspace}
\newcommand{\ufpralpr}{UFPR-ALPR\xspace}

%% file: 0-aux/variables.tex
\newcommand{\supplementary}{\url{https://github.com/valfride/lpr-rsr/}}

%% file: 0-aux/abstract.tex
\ifieee
\vspace{-3.5mm}
\else
\fi
\begin{abstract}
The \gls*{lpr} field has made impressive advances in the last decade due to novel deep learning approaches combined with the increased availability of training data.
However, it still has some open issues, especially when the data come from \gls*{lr} and low-quality images/ videos, as in surveillance systems.
This work focuses on \gls*{lp} reconstruction in \gls*{lr} and low-quality images.
We present a \gls*{sisr} approach that extends the attention/transformer module concept by exploiting the PixelShuffle layers capabilities and that has an improved loss function based on \gls*{lpr} predictions.
For training the proposed architecture, we use synthetic images generated by applying heavy Gaussian noise in terms of \gls*{ssim} to the original \gls*{hr} images.
In our experiments, the proposed method outperformed the baselines both quantitatively and qualitatively.
The datasets we created for this work are publicly available to the research community at \textit{\supplementary}.

\end{abstract}

%% file: 1-Introduction/Introduction.tex
\section{Introduction}

\glsresetall

Image quality improvement is a recurrent yet challenging topic in computer vision due to its complexity and high practical value in many surveillance applications, for example, \gls*{lp}, face, and object recognition.
In this regard, \gls*{sr}~\cite{wang2021deep} has been an important research subject within the last few decades due to its capacity to retrieve subtleties and textures from \gls*{lr} images and generate their \gls*{hr} counterparts.
Moreover, the storage of \gls*{hr} images as \gls*{lr} versions and the ability to recover them when necessary is desirable~\cite{gabriele2021perspective, liu2022blind}.

\gls{sr} is typically divided into \gls*{sisr}, multi-image super-resolution, and video super-resolution.
Here, we focus on~\gls*{sisr}.
There are two main reasons why \gls*{sr} is still considered a challenging research topic.
First, it is an inherently ill-posed problem since one \gls*{lr} image may have multiple plausible \gls*{hr} reconstructions~\cite{nasrollahi2014superresolution}.
Second, increasing the upscaling factor also increases the complexity of the problem~\cite{wang2021deep}.

With the advancement of deep learning methods and their chain of success in computer vision problems, the spread of \glspl*{cnn} to deal with \gls*{sr} applications can be easily noticed~\cite{lucas2019generative,mehri2021mprnet,liu2022blind}.
Although significant advances have been made, most of the approaches to solving the \gls*{sr} problem are based on very deep architectures, which increase general computation operations, and focus only on achieving a higher \gls*{psnr} and \gls*{ssim}, disregarding contextual information of the application at hand.
In the \gls*{lpr} context, we consider this is not the best way to handle the problem as one approach may create very realistic images even if it fails to differentiate similar characters (e.g., ‘Q’ and ‘O’, ‘1’ and ‘I’, among~others).

In this work, \gls*{ps} layers are employed in a new perspective to improve \gls*{lp} super-resolution by extending the \gls*{mprnet}~\cite{mehri2021mprnet} attention module.
In addition, shallow features from input images are extracted by squeezing and expanding with an auto-encoder built with \gls*{ps} and \gls*{pu} layers.
Lastly, considering our target application (i.e., \gls*{lpr}), we propose a loss function that also considers the predictions returned by an \gls*{ocr} model~\cite{goncalves2018realtime} on the reconstructed~images.

In summary, the main contributions of our work are:
\begin{itemize}
    
    \item An extension of the attention mechanism from \gls*{mprnet} that uses \gls*{ps} layers for channels reorganization~\cite{shi2016realtime};

    \item A new loss function that incorporates both~\gls*{lpr} predictions and quality metrics in its~formulation;

    \item The datasets we built for this research (images degraded at different \gls*{ssim} intervals) are publicly~available\footnote{Access is granted \emph{upon request}, i.e., interested parties must register by filling out a registration form and agreeing to the dataset's terms of use.}.

\end{itemize}

%% file: 2-RelatedWork/RelatedWork.tex
\section{Related Work}

This section briefly describes related work.
An overview of \gls*{sisr} approaches is given in \cref{sisr}, while deep learning methods for \gls*{lp} super-resolution are described in \cref{srOnLp}.

\subsection{Single-Image Super-Resolution}

\label{sisr}

Dong et al.~\cite{dong2016image} proposed one of the first deep learning-based methods, called \gls*{srcnn}, to tackle the SR ill-posed characteristic.
It showed to be faster and quantitatively better in restoration capabilities than previous example-based methods, with fewer pre- or post-processing~steps.

Despite its success, \gls*{srcnn} receives pre-upsampled \gls*{lr} images generated through interpolation methods, drastically increasing computational complexity without aggregating vital information to further image restoration~\cite{7527621, wang2015deep}.
Later, Dong et al.~\cite{dong2016accelerating} and Shi et al.~\cite{shi2016realtime} explored the upsample process near the end of the network as part of the architecture pipeline.
This strategy resulted in an expressive reduction of run time, parameters, and computational cost.
In~\cite{dong2014learning}, Dong et al. observed that bicubic interpolation is also a convolutional operation, so it can be formulated as a convolutional layer.

The importance of learnable upscaling was highlighted by Shi et al.~\cite{shi2016realtime}.
They designed specialized convolution layers to learn upscaling filters, making it possible to learn more complex mappings from \gls*{lr} to \gls*{hr} images with increased performance compared to fixed-size interpolation~methods.

The presence of ``\textit{dead}'' filters was observed both in~\cite{dong2014learning} and~\cite{yang2013fast}.
These dead filters may be seen as the network alone trying to ``\textit{discover}'' what are the essential features from the input, resulting in a lack of learning and performance.

Accordingly, to better influence the network into allocating available computer resources to the most informative aspects of the input images, Zhang et al.~\cite{zhang2018image} introduced the concept of first-order statistic channel attention mechanism for image reconstruction that uses only information across inner-channel features to image reconstruction.
Afterward, Dai et al.~\cite{dai2019second} proposed the second-order version of the attention model to explore more meaningful features expression.

More recently, motivated by these works, Mehri et al.~\cite{mehri2021mprnet} presented the \gls*{tfam} to exploit essential information --~considering both inner-channel and spatial features~-- to boost the network's performance.
Their results showed superior or competitive performance compared to multiple baselines~\cite{dong2014learning, luo2020latticenet, zhang2018residual}. \gls*{mprnet} generated \gls*{sr} images with textures similar to the original \gls*{hr} images since it fully uses the abstract features within the \gls*{lr}~input.

\subsection{Super-Resolution for License Plate Recognition}
\label{srOnLp}

The main goal of an \gls*{lpr} system is to extract the \gls*{lp} information from an image or sequence of images~\cite{zeni2020weakly,laroca2022cross}.
These systems have been extensively researched due to their practical applications in security tasks such as traffic law enforcement, monitoring private areas, and criminal investigations~\cite{weihong2020research}.

Although impressive results have been reported in recent years in the \gls*{lpr} context~\cite{laroca2021efficient,wang2022rethinking,silva2022flexible}, the datasets where the proposed models are being evaluated are mostly composed of \gls*{hr} images where all \gls*{lp} characters are pretty legible.
In most surveillance scenarios, this is not in line with everyday~reality.

The quality of \gls*{lp} images is intrinsically related to various factors, such as camera distance, motion blur, lighting conditions, and image compression technique used for storage~\cite{goncalves2019multitask}.
While commercial \gls*{lpr} systems capture sharp images using \textit{global shutter} cameras, cheaper cameras using \textit{rolling shutter} technology are typically employed in surveillance systems, often resulting in blurry images~\cite{liang2008analysis} with illegible~\glspl*{lp}.

\begin{figure*}[!htb]
\centering
    \includegraphics[width=0.75\linewidth]{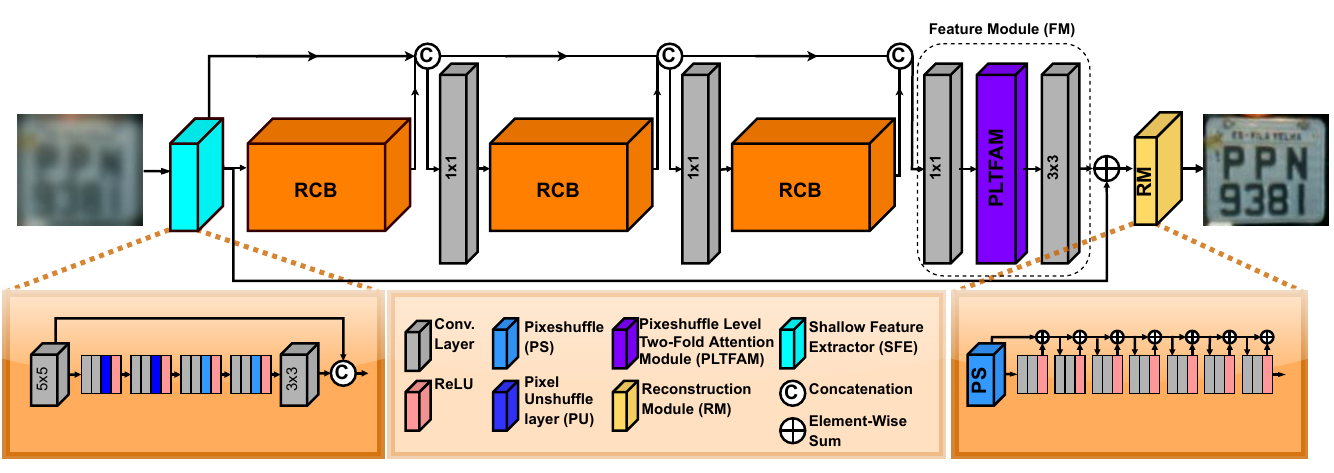}
    
    \vspace{-4mm}
    
    \caption{The proposed architecture; acronyms are described in the internal legend. To the network was introduced an autoencoder composed of \gls{pu} and \gls{ps} layers for squeeze and expansion, respectively, aiming to exclude less relevant features. \gls*{pltfam} replaced the original \gls{tfam} modules along the~network.}
    \label{fig:nnarch}
\end{figure*}

While the first works to combine the idea of SR and LP recognition date from the 2000s~\cite{4220667,yuan2008fast}, this research area has received
increasing attention in recent years given the rise of deep learning.
Considering space limitations, the remainder of this section describes works published in recent years.

Svodoba et al.~\cite{svoboda2016cnn} demonstrated that \glspl*{cnn} trained on artificially generated blurry images provide superior quality enhancement on images with motion blur compared to traditional blind deconvolution methods. 
However, as they trained their model for a specific range of motion blur lengths and directions, the reconstruction quality degrades considerably for blurs outside the range of the blurs the network was trained~for.

Lin et al.~\cite{lin2021license} exploited the high capability of a \gls*{gan} for \gls*{lp} reconstruction. 
They reported promising results; nevertheless, their experiments were carried out on just $100$ images.
Moreover, their approach was compared with other methods only in terms of \gls*{psnr} and \gls*{ssim}, without exploring \gls*{lp} recognition at~all.

In the same direction, Hamdi et al.~\cite{hamdi2021new} concatenated two \gls*{gan} models for this task.
The first was used for denoising and deblurring, while the second was applied to super-resolution.
The authors compared their method with three baselines, but only in terms of \gls*{psnr} and \gls*{ssim} as well.
Interestingly, after analyzing the results, they acknowledged that higher \gls*{psnr} and \gls*{ssim} do not necessarily mean better~reconstruction.

Lee et al.~\cite{lee2020super} observed that previous \gls*{sr} approaches did not take character recognition into account. 
Thus, they designed a \gls*{gan}-based model that relies on a perceptual loss composed of intermediate features extracted by a scene text recognition model.
While their method reported better results than the same \gls*{gan}-based model trained with the original perceptual loss, their dataset was not made available and the degradation method they used was not detailed as~well.

While the final goal in enhancing \gls*{lp} images is to improve the recognition results, in current works (except~\cite{lee2020super}) the quality of the reconstructed images was evaluated either qualitatively or based on the \gls*{psnr} and \gls*{ssim} metrics, which are known not to correlate well with human assessment of visual
quality~\cite{johnson2016perceptual,zhang2018unreasonable}.
Also, in most related works the experiments were conducted exclusively on private datasets~\cite{svoboda2016cnn,lee2020super,hamdi2021new}.

%% file: 3-Methodology/Methodology.tex
\section{Methodology}
This section describes the proposed approach.
We first detail how we extend the \gls{mprnet} architecture proposed by Mehri et al.~\cite{mehri2021mprnet}.
Then, we present our improved loss~function.

\subsection{Network Architecture Modifications}

Inspired by the work of Mehri et al.~\cite{mehri2021mprnet}, the proposed network architecture is presented in \cref{fig:nnarch}.
It consists of four different modules, namely, \gls*{sfe}; \gls*{rcb}; \gls*{fm}; and reconstruction module, which combines the output of the \gls{fm} module with a long-skip connection from the end of the \gls*{sfe} module.
We highlight our changes in the followings.

\gls*{sfe} block internal design consists mainly of a \gls*{psfe} with a $5\times5$ kernel and a convolution layer followed by an autoencoder with \gls*{pu} and \gls*{ps} layers in its composition instead of classic pooling and upscale operations. Finally, the autoencoder output is combined with a skip connection from \gls{psfe}.
The main idea %
of this design is to learn and emphasize the most important characteristics by squeezing~(\gls*{pu}) and expanding~(\gls*{ps}) to the original dimensions with an aggregation $3\times3$ convolution layer at the end.
Less informative features are not lost thanks to the skip connection.

\begin{figure}[!t]
    \centering
    \captionsetup[subfigure]{captionskip=-0.25pt}
    
    \resizebox{0.9\linewidth}{!}{
    \subfloat[\phantom{-i}\label{subfig-1:dummy}]{%
        \includegraphics[width=0.43\linewidth]{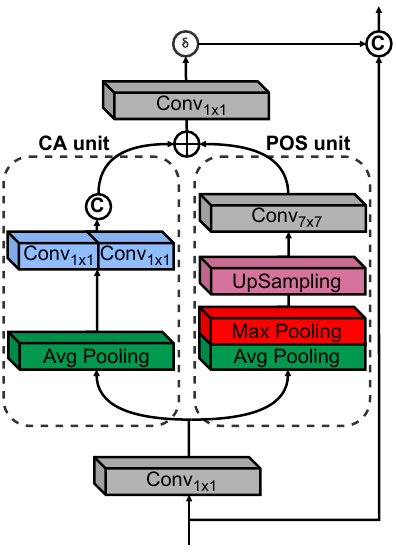}
    } \qquad
    \subfloat[\phantom{--,}\label{subfig-2:dummy}]{%
        \includegraphics[width=0.43\linewidth]{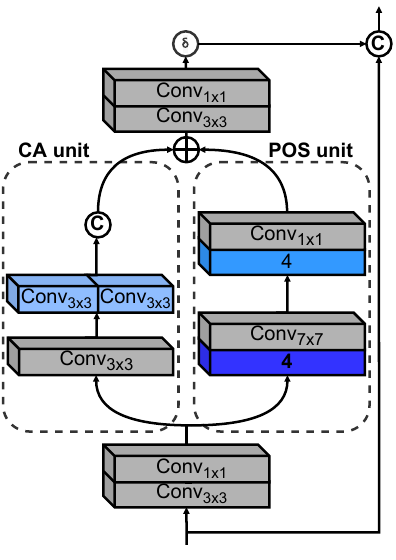}
    }
    }
    
    \vspace{-1.25mm}
    
    \caption{Comparative illustration of the (a)~two-fold attention module in MPRNet~\cite{mehri2021mprnet} and (b)~Pixelshuffle two-fold attention module~(ours). 
}
    \label{fig:pltfam}
\end{figure}

Fig.~\ref{fig:pltfam} highlights the improvements we made from \gls*{mprnet}'s \gls*{tfam}~\cite{mehri2021mprnet} to produce our proposed \gls*{pltfam} (the purple block in \cref{fig:nnarch}),
which was built following the insights:
(i) images are composed of inter-channel relationships since each channel contributes with unique characteristics to compose the final image.
Therefore, the extraction of such key features is essential;
(ii) learning positional information of the key characteristics from each channel that compose the inter-channel relationships is required; and (iii) as stated by Shi et al.~\cite{shi2016realtime} and Dong et al.~\cite{dong2014learning}, downscale and upscale layers rely on  Translational Invariance (e.g., \textit{MaxPool} and \textit{AvgPool}) and interpolation (e.g., bicubic) general techniques, thus, both layers are not able to learn a custom process for different tasks.

\gls*{ca} is concerned about summarizing inter-channel relationship features and excluding less relevant ones. 
For such an aim, it starts with a \gls*{ps} layer to better explore the most important features by optimally rearranging pixels to emphasize meaningful features.
Later, two convolution layers --~working side by side~-- receive half the input followed by the concatenation of the outputs to leverage the previous~operation.

The positional unit learns where the significant features within the image are located.
It decreases and increases the input by the same scale factor with \gls*{ps} and \gls*{pu} layers in sequence.
As a result of this process, the positions of the relevant inter-channel relationship features are highlighted. 
Lastly, the results from both \gls*{ca} and \gls*{pos} proceed to an element-wise sum, followed by a $3\times3$ and $1\times1$ kernel convolution layers and a sigmoid function to generate the final attention mask, which aggregates all the relevant information extracted by \gls*{ca} and \gls*{pos} units.

The \glspl*{rcb} had their original \gls*{tfam} replaced by the \gls*{pltfam}, but its main structure remained the same adopted in~\cite{mehri2021mprnet}.

Returning our attention to \cref{fig:nnarch}, the reconstruction module was added as an output block for better aggregating fine details.
It comprises one \gls*{ps} for pixel reorganization, followed by seven recurrent blocks with two $3\times3$ kernel size convolution layers and ReLU activation function.
A sigmoid activation function was added at the end of the reconstruction module to limit the values to $[0, 1]$ and thus smooth the~output.

\subsection{Perceptual Loss}

Considering that the final goal of an \gls*{lpr} system is to achieve recognition rates as high as possible, we propose the following perceptual loss:
\begin{equation}
    PL=\frac{1}{n}\sum_{i=0}^{n}(H_i-S_i)^{2}(1+\alpha \cdot D(H_i, S_i)),
\end{equation}
\noindent where the details term $D(H_i, S_i)$ for the $i$-th high-resolution image $H_i$ and its respective super-resolution $S_i$ stands for:
\begin{equation}
D(H_i, S_i) = Lev(H_{i},S_{i})/7 + (1-SSIM(H_{i},S_{i})).
\end{equation}

Here, the loss is weighted by the trade-off between visual quality and recognition rate using $D(H_i, S_i)$. 
This task is accomplished by measuring the Levenshtein distance (also known
as edit distance) and comparing the \gls*{ssim} score between the \gls*{sr} and \gls*{hr} images. 
The resulting value is scaled to the same magnitude as the squared error between the \gls*{sr} and \gls*{hr} images using $\alpha$, thus avoiding dominance from either~term. 

Notably, any \gls*{ocr} model can be employed for \gls*{lp} recognition in this loss function.
Such flexibility is attractive since we can straightforwardly use novel models as they are introduced.
In this work, we explored the multi-task model proposed by Gon\c{c}alves et al.~\cite{goncalves2018realtime}, as it was specifically designed for recognizing Brazilian \glspl*{lp} and has high~efficiency.

%% file: 4-Experiments/Experiments.tex
\section{Experiments}

This section describes the experiments performed to validate the proposed approach.
We first present our experimental setup and then report the results~obtained.

\subsection{Setup}
\label{subsec:setup}
    
We conducted our experiments using PyTorch on a computer with an AMD Ryzen $9$ $5950$X CPU, $128$~GB of RAM, and an NVIDIA Quadro RTX~$8000$ GPU~($48$~GB).

\subsubsection{Dataset}

we performed our experiments on \gls*{lp} images extracted from the \rodosolalpr dataset~\cite{laroca2022cross}.
This dataset comprises $20{,}000$ images taken by static cameras at pay rolls located in the Brazilian state of Esp\'{\i}rito Santo.

Among the $20{,}000$ images, there are $5{,}000$ images of each of the following combinations of vehicle type and \gls*{lp} layout: (i)~cars with Brazilian \glspl*{lp}, (ii)~cars with Mercosur \glspl*{lp}, (iii)~motorcycles with Brazilian \glspl*{lp}, and (iv)~motorcycles with Mercosur \glspl*{lp}\footnote{Following previous works~\cite{goncalves2018realtime,laroca2021efficient,silva2022flexible}, we refer to ``Brazilian'' as the layout used in Brazil before the adoption of the Mercosur~layout.}. 
While all Brazilian \glspl*{lp} consist of three letters followed by four digits, the initial pattern adopted in Brazil for Mercosur \glspl*{lp} consists of three letters, one digit, one letter and two digits, in that order~\cite{laroca2022cross} (this is the pattern adopted on all Mercosur \glspl*{lp} in the \rodosolalpr dataset).

\begin{figure}[!tb]
    \centering
    
    \resizebox{0.825\linewidth}{!}{
    \includegraphics[width=0.31\linewidth]{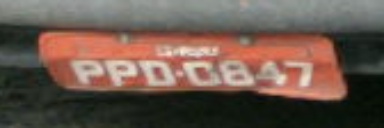} \hspace{-1.5mm}
    \includegraphics[width=0.31\linewidth]{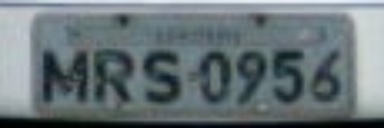} \hspace{-1.5mm}
    \includegraphics[width=0.31\linewidth]{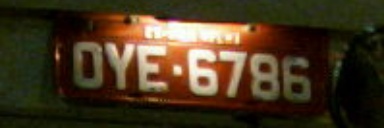}
    }
    
    \vspace{0.65mm}
    
    \resizebox{0.825\linewidth}{!}{
    \includegraphics[width=0.31\linewidth]{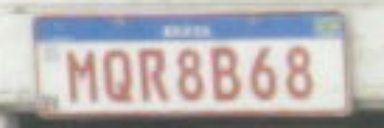} \hspace{-1.5mm}
    \includegraphics[width=0.31\linewidth]{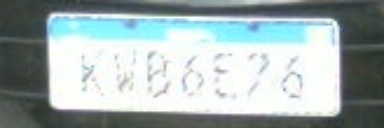} \hspace{-1.5mm}
    \includegraphics[width=0.31\linewidth]{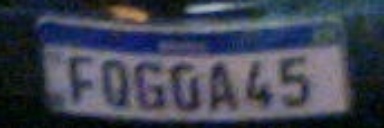}
    }
    
    \vspace{0.6mm}
    
    \resizebox{0.825\linewidth}{!}{
    \includegraphics[width=0.31\linewidth]{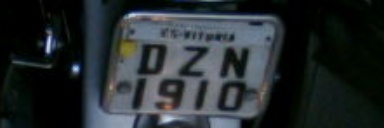} \hspace{-1.5mm}
    \includegraphics[width=0.31\linewidth]{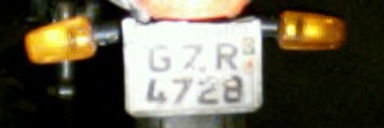} \hspace{-1.5mm}
    \includegraphics[width=0.31\linewidth]{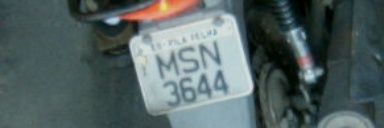}
    }
    
    \vspace{0.6mm}
    
    \resizebox{0.825\linewidth}{!}{
    \includegraphics[width=0.31\linewidth]{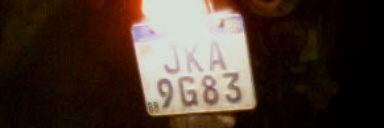} \hspace{-1.5mm}
    \includegraphics[width=0.31\linewidth]{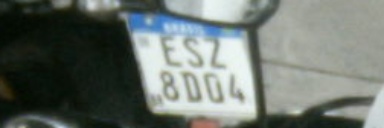} \hspace{-1.5mm}
    \includegraphics[width=0.31\linewidth]{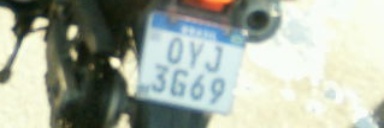}
    }
    
    \vspace{-2.75mm}
    
    \caption{Examples of cropped \glspl*{lp} from the \rodosolalpr dataset~\cite{laroca2022cross}.}
    \label{fig:samples-rodosol}
\end{figure}

As far as we know, \rodosolalpr is the largest public dataset in terms of both Brazilian and Mercosur \glspl*{lp}.
\cref{fig:samples-rodosol} shows some \glspl*{lp} cropped from the \rodosolalpr~dataset.
Observe the diversity of this dataset regarding several factors such as \gls*{lp} colors, lighting conditions, and character fonts.

The \gls*{hr} images used in our experiments were generated as follows.
For each image from the \rodosolalpr dataset, we first cropped the \gls*{lp} region using the annotations provided by the authors.
Afterward, considering the scope of this work, we used the same annotations to rectify each \gls*{lp} image so that it becomes more horizontal, tightly bounded, and easier to recognize~\cite{laroca2021towards}.
The rectified image is the \gls*{hr}~image.

For each \gls*{hr} image, we generate multiple \gls*{lr} images.
Inspired by~\cite{zhang2018residual}, we simulated the effect of an optical system with a lower resolution by iteratively applying random Gaussian noise to each \gls*{hr} image until the desired degradation level for a given \gls*{lr} image was reached.
Intuitively, we measure the level of degradation of an \gls*{lr} image considering the \gls*{ssim} score between it and the respective \gls*{hr} image.

\subsubsection{Training}
in the training stage, the \gls*{lr} and \gls{hr} images were first padded to preserve the aspect ratio and then resized to $120 \times 60$ pixels with no upsample step on the subsequent \gls*{sr} outputs.
We created four subsets at $]0.00,0.10]$, $]0.10,0.25]$, $]0.25,0.50]$ and $]0.50,0.75]$ \gls{ssim} intervals each with $8{,}000$ and $4{,}000$ images for training and validation, respectively.
We also trained the proposed network and MPRNet~\cite{mehri2021mprnet} using the union of all subsets (i.e., $]0.00,0.75]$ \gls*{ssim}~interval).

We used the Adam optimizer with an underlying learning rate of $10$\textsuperscript{-$4$}, which decreases by a factor of $0.8$ (up to $10$\textsuperscript{-$7$}) when no improvement in the loss function is observed.
The training stage stops after $5$ epochs without loss~improvement.

\subsubsection{Testing}

the proposed models were tested on the remaining images: $8{,}000$ images from each of the $]0.00,0.10]$, $]0.10,0.25]$, $]0.25,0.50]$, and $]0.50,0.75]$ subsets.
For the model and baseline trained with the $]0.00,0.75]$ \gls{ssim} interval, the above subsets were combined.
For each experiment, we report the number of correctly recognized \glspl*{lp} divided by the number of \glspl*{lp} in the test set.
A correctly recognized \gls*{lp} means that all characters on the \gls*{lp} were correctly recognized.

\begin{figure}[!htb]
    \centering
    
    \resizebox{0.85\linewidth}{!}{
    \includegraphics[width=0.31\linewidth]{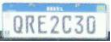} \hspace{-1.25mm}
    \includegraphics[width=0.31\linewidth]{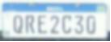} \hspace{-1.25mm}
    \includegraphics[width=0.31\linewidth]{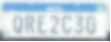} \hspace{-1.25mm}
    \includegraphics[width=0.31\linewidth]{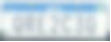} \hspace{-1.25mm}
    \includegraphics[width=0.31\linewidth]{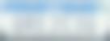} 
    }
    
    \vspace{0.6mm}
    
    \resizebox{0.85\linewidth}{!}{
    \includegraphics[width=0.31\linewidth]{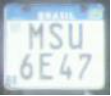} \hspace{-1.25mm}
    \includegraphics[width=0.31\linewidth]{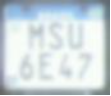} \hspace{-1.25mm}
    \includegraphics[width=0.31\linewidth]{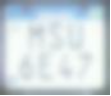} \hspace{-1.25mm}
    \includegraphics[width=0.31\linewidth]{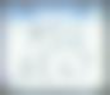} \hspace{-1.25mm}
    \includegraphics[width=0.31\linewidth]{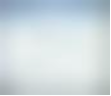} 
    }
    
    \vspace{0.6mm}

    \resizebox{0.85\linewidth}{!}{
    \includegraphics[width=0.31\linewidth]{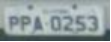} \hspace{-1.25mm}
    \includegraphics[width=0.31\linewidth]{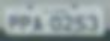} \hspace{-1.25mm}
    \includegraphics[width=0.31\linewidth]{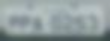} \hspace{-1.25mm}
    \includegraphics[width=0.31\linewidth]{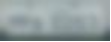} \hspace{-1.25mm}
    \includegraphics[width=0.31\linewidth]{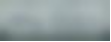} 
    }
    
    \vspace{0.6mm}
    
    \resizebox{0.85\linewidth}{!}{
    \includegraphics[width=0.31\linewidth]{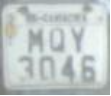} \hspace{-1.25mm}
    \includegraphics[width=0.31\linewidth]{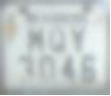} \hspace{-1.25mm}
    \includegraphics[width=0.31\linewidth]{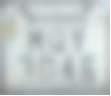} \hspace{-1.25mm}
    \includegraphics[width=0.31\linewidth]{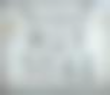} \hspace{-1.25mm}
    \includegraphics[width=0.31\linewidth]{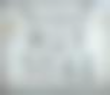} 
    }
    
    \vspace{-2.5mm}
    
    \caption{Representative samples of the subsets used in our experiments. From left to right: $1$ (original image), $0.75$, $0.5$, $0.25$ and $0.1$ \gls*{ssim} scores.}
    \label{fig:SSIM_examples}
\end{figure}

\subsection{Results}
\label{subsec:results}

To illustrate how challenging is the problem at hand, 
the first section of \cref{tab:ExperimentalResults} presents the recognition rates for the \gls*{hr} image and the respective \gls*{lr} images degraded by recursive Gaussian noise aiming for different \gls*{ssim} interval scores, i.e., $]0.00,0.10]$, $]0.10,0.25]$, $]0.25,0.50]$ and $]0.50, 0.75]$.
\cref{fig:SSIM_examples} shows some representative examples of the images used in our experiments.
Note that the \gls*{lp} characters are visually hard to distinguish in images with \gls*{ssim} scores lower than $0.50$.
Observe that the recognition rates achieved on motorcycle \glspl*{lp} are higher than those achieved on car \glspl*{lp} in less degraded images ($]0.25,0.50]$ and $]0.50, 0.75]$), but not in considerably degraded images ($]0.00,0.10]$ and $]0.10,0.25]$).
This occurs because motorcycle \glspl*{lp} are generally smaller in size (having less space
between the characters) and are often tilted~\cite{laroca2022cross}. Consequently, through the degradation process, the characters are mixed together, affecting the recognition~performance.

\begin{table}[!htb]
\centering
\setlength{\tabcolsep}{1.75pt}
\renewcommand{\arraystretch}{1.05}
\caption{Recognition rates (\%) achieved in our experiments.}

\vspace{-3mm}

\resizebox{0.925\linewidth}{!}{
\begin{tabular}{lccccccccccccccc}
\toprule
\multicolumn{1}{c}{\multirow{2}{*}{\gls*{ssim}}} & & & \multicolumn{3}{c}{Cars} & & & \multicolumn{3}{c}{Motorcycles} & & & \multicolumn{3}{c}{~Cars \& Motor.} \\
                & & & All & $\leq6$ & $\leq5$ &  & & All  & $\leq6$ & $\leq5$& & & All & $\leq6$ & $\leq5$ \\
\midrule                
\multicolumn{16}{l}{No super-resolution}\\
\midrule
\indentresults \gls*{hr}      & & & 90.9 & 97.5 & 98.9 & & & 95.2 & 99.5 & 99.9 & & & 92.8 & 98.4 & 99.4 \\[0.75pt] \cdashline{1-16} \\[-6.75pt]
\indentresults $]0.50,0.75]$   & & & 88.7 & 96.6 & 98.6 & & & 93.6 & 99.2 & 99.8 & & & 90.9 & 97.7 & 99.1\\
\indentresults $]0.25,0.50]$   & & & 73.5 & 88.3 & 94.3 & & & 81.7 & 95.3 & 98.3 & & & 77.2 & 91.4 & 96.1\\
\indentresults $]0.10,0.25]$   & & & 18.4 & 35.9 & 53.4 & & & 17.7 & 39.1 & 59.0 & & & 18.1 & 37.3 & 55.9\\
\indentresults $]0.00,0.10]$      & & & \phantom{0}0.3 & \phantom{0}1.3 &  \phantom{0}4.8 & & & \phantom{0}0.1 &   \phantom{0}0.9 & \phantom{0}3.7 & & & \phantom{0}0.2 &  \phantom{0}1.1 & \phantom{0}4.3\\
\midrule
\multicolumn{16}{l}{Proposed model trained with $]0.00,0.10]$ \gls*{ssim} images}\\
\midrule
\indentresults $]0.50,0.75]$    & & & 36.7 & 62.0 & 78.5 & & & 29.7 & 52.0 & 68.5 & & & 33.5 & 57.5 & 74.0 \\
\indentresults $]0.25,0.50]$    & & & 31.2 & 55.7 & 72.6 & & & 24.8 & 46.1 & 62.2 & & & 28.3 & 51.4 & 68.0 \\
\indentresults $]0.10,0.25]$    & & & 26.9 & 50.2 & 66.2 & & & 21.7 & 43.7 & 62.0 & & & 24.5 & 47.3 & 64.3 \\
\indentresults $]0.00,0.10]$    & & & 17.4 & 35.1 & 50.5 & & & \phantom{0}8.5 & 20.7 & 36.1 & & & 13.4 & 28.7 & 44.0 \\
\midrule
\multicolumn{16}{l}{Proposed model trained with $]0.10,0.25]$ SSIM images}\\
\midrule
\indentresults $]0.50,0.75]$   & & & 56.0 & 74.7 & 85.6 & & & 55.0 & 74.8 & 84.7 & & & 55.6 & 74.7 & 85.2 \\
\indentresults $]0.25,0.50]$   & & & 56.4 & 75.6 & 85.3 & & & 59.2 & 78.5 & 88.6 & & & 57.7 & 76.9 & 86.8 \\
\indentresults $]0.10,0.25]$  & & & 60.3 & 79.1 & 87.4 & & & 52.6  & 76.7 & 88.6 & & & 56.8 & 78.0 & 88.0 \\

\indentresults $]0.00,0.10]$   & & & 13.4 & 27.3 & 40.0 & & & \phantom{0}5.9 & 14.8 & 25.5 & & & 10.0 & 21.7 & 33.5 \\
\midrule
\multicolumn{16}{l}{Proposed model trained with $]0.00, 0.75]$ SSIM images}\\
\midrule
\indentresults $]0.50,0.75]$   & & & 90.6 & 97.4 & 98.9 & & & 93.8 & 98.9 & 99.7 & & & 92.0 & 98.0 & 99.3 \\
\indentresults $]0.25,0.50]$   & & & 87.3 & 96.1 & 98.5 & & & 91.2 & 98.0 & 99.5 & & & 89.0 & 96.9 & 98.9 \\
\indentresults $]0.10,0.25]$   & & & 69.3 & 85.7 & 92.9 & & & 65.5 & 85.5 & 93.6 & & & 67.6 & 85.6 & 93.2 \\
\indentresults $]0.00,0.10]$   & & & 32.1 & 51.1 & 65.1 & & & 13.9 & 30.9 & 47.6 & & & 23.9 & 42.1 & 57.3 \\[1.25pt] %
\midrule
\multicolumn{16}{l}{Proposed model \& baselines trained and tested with $]0, 0.75]$ \gls*{ssim} images}\\
\midrule
\indentresults \textbf{Proposed}    & & & \textbf{69.8} & \textbf{82.6} & \textbf{88.9} & & & \textbf{66.1} & \textbf{78.3} & \textbf{85.1} & & & \textbf{68.1} & \textbf{80.7} & \textbf{87.2} \\
\indentresults LR-LPR (no SR)~\cite{goncalves2019multitask} & & & 61.4 & 78.0 & 86.5 & & & 47.0 & 68.8 & 80.4 & & & 54.9 & 73.9 & 83.7 \\ %
\indentresults MPRNet~\cite{mehri2021mprnet}   & & & 48.2 & 66.1 & 75.7 & & & 50.0 & 65.0 & 74.6 & & & 49.0  & 65.6 & 75.2 \\ 
\midrule
\multicolumn{16}{l}{Average PSNR~(dB) and SSIM for tests with $]0, 0.75]$ SSIM images}\\ \midrule
  & & & & &  PSNR  & & & & &  SSIM & & & & &   \\
\indentresults \textbf{Proposed} & & & & &  \textbf{26.4}  & & & & &  \textbf{0.89} & & & & & \\
\indentresults MPRNet~\cite{mehri2021mprnet}   & & & & & 19.7 & & & & & 0.79 & & & & & \\
\bottomrule
\end{tabular}
}
\label{tab:ExperimentalResults}
\end{table}

In the second and third sections of \cref{tab:ExperimentalResults}, we present the recognition rates obtained in \gls*{sr} images generated by the proposed network when trained on \gls*{lr} images with \gls{ssim} in the $]0.00,0.10]$ and $]0.10,0.25]$ intervals, respectively.
As expected, considerably better recognition rates are achieved in \gls*{sr} images generated by the model trained on images from the same \gls*{ssim} interval.
However, the \gls*{ocr} network did not perform well on \gls*{sr} images generated by models trained on images in different \gls*{ssim} intervals, especially those with better quality indices, i.e., $]0.25,0.50]$ and $]0.50,0.75]$.
Based on the recognition results shown in the fourth section of \cref{tab:ExperimentalResults}, we can state that the \gls*{sr} model trained on images with \gls*{ssim} in the $]0.00, 0.75]$ interval generalized much better than models trained exclusively on images from $]0.00,0.10]$ or~$]0.10,0.25]$.

In the next-to-last section of Table~\ref{tab:ExperimentalResults}, we compare the proposed architecture with MPRNet~\cite{mehri2021mprnet}\footnote{\hspace{0.2mm}We implemented \gls*{mprnet} ourselves, as its code has not been made public.}.
The \gls*{ocr} network~\cite{goncalves2018realtime} achieved considerably better results on images reconstructed by our \gls*{sr} model than \gls*{mprnet}.
We believe this is due to the capabilities of \gls*{ps} and \gls*{pu} in learning the best way to scale and reorganize channels within the image.
We also report the recognition rates obtained by the \gls*{ocr} model proposed in~\cite{goncalves2019multitask}, henceforth called LR-LPR, as it was specifically designed for recognizing low-resolution \glspl*{lp}.
Note that we trained and tested LR-LPR as in~\cite{goncalves2019multitask}, that is, using the degraded images (in the $]0.00, 0.75]$ \gls*{ssim} interval) and not reconstructed ones.
Although it achieved better recognition rates than we expected, the proposed approach (i.e., first reconstructing the degraded \gls*{lp} images using our \gls*{sr} model and then feeding the resulting images into an \gls*{ocr} network trained on \gls*{hr} \glspl*{lp}) still achieved significantly superior~results.

As can be seen at the bottom of \cref{tab:ExperimentalResults}, the proposed architecture achieved significantly better results than \gls*{mprnet}~\cite{mehri2021mprnet} also considering the \gls*{ssim} and \gls*{psnr}~(dB) metrics, reaching $0.89$ and $26.4$~dB against $0.79$ and $19.7$~dB,~respectively.

\begin{figure}[!htb]
    \captionsetup[subfigure]{labelformat=empty,position=top,captionskip=0.75pt,justification=centering}

    \vspace{0.7mm}
    
    \centering
    \resizebox{0.825\linewidth}{!}{
    \subfloat[LR (Input)]{
    \includegraphics[width=0.2\linewidth]{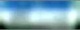}
    } \hspace{-2.25mm}
    \subfloat[\gls*{mprnet}~\cite{mehri2021mprnet}]{
    \includegraphics[width=0.2\linewidth]{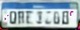}
    } \hspace{-2.25mm}
    \subfloat[Proposed]{
    \includegraphics[width=0.2\linewidth]{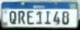}
    } \hspace{-2.25mm}
    \subfloat[HR (GT)]{
    \includegraphics[width=0.2\linewidth]{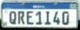} 
    } \,
    }
    
    \vspace{-2mm}
    
    \resizebox{0.825\linewidth}{!}{
    \subfloat[]{
    \includegraphics[width=0.2\linewidth]{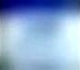}
    } \hspace{-2.25mm}
    \subfloat[]{
    \includegraphics[width=0.2\linewidth]{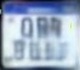} 
    } \hspace{-2.25mm}
    \subfloat[]{
    \includegraphics[width=0.2\linewidth]{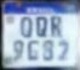} 
    } \hspace{-2.25mm}
    \subfloat[]{
    \includegraphics[width=0.2\linewidth]{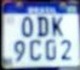} 
    } \,
    }
    
    \vspace{-2mm}

    \resizebox{0.825\linewidth}{!}{
    \subfloat[]{
    \includegraphics[width=0.2\linewidth]{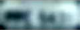}
    } \hspace{-2.25mm}
    \subfloat[]{
    \includegraphics[width=0.2\linewidth]{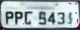}
    } \hspace{-2.25mm}
    \subfloat[]{
    \includegraphics[width=0.2\linewidth]{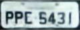}
    } \hspace{-2.25mm}
    \subfloat[]{
    \includegraphics[width=0.2\linewidth]{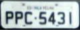} 
    } \,
    } 
    
    \vspace{-2mm}

    \resizebox{0.825\linewidth}{!}{
    \subfloat[]{
    \includegraphics[width=0.2\linewidth]{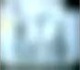}
    } \hspace{-2.25mm}
    \subfloat[]{
    \includegraphics[width=0.2\linewidth]{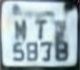}
    } \hspace{-2.25mm}
    \subfloat[]{
    \includegraphics[width=0.2\linewidth]{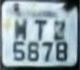}
    } \hspace{-2.25mm}
    \subfloat[]{
    \includegraphics[width=0.2\linewidth]{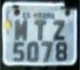} 
    } \,
    }

    \caption{Qualitative results on models trained with ]0, 0.75] SSIM images.}%
    \label{fig:Qresults}
\end{figure}

Finally, \cref{fig:Qresults} shows four \gls*{lr} images and the respective \gls*{sr} images generated by our architecture and by \gls*{mprnet}~\cite{mehri2021mprnet}.
The original image is also shown for better comparison.

The \gls*{lr} images shown in \cref{fig:Qresults} are in the range of $]0, 10]$ \gls*{ssim}; thus, the \glspl*{lp} are heavily degraded with little to no visible characters, representing a challenging reconstruction scenario for any \gls*{sr} model.
In general, the proposed architecture performs better than \gls*{mprnet}~\cite{mehri2021mprnet} on perceptual reconstruction.
Observe that \gls{mprnet} tends to produce blurred edges, which in most cases mixes the characters with each other or with the \gls*{lp} background; this is very noticeable in the second row's image, where most \gls*{lp} characters are considerably~blurred. 
In contrast, our model generates sharper character edges, which promotes better differentiation from the \gls*{lp} background. Also, its character reconstruction is not inconsistent, with incomplete lines or missing characters.
When the \gls*{sr} model does not know which character to represent, it tends to hallucinate for the most congruent ones with respect to the \gls*{lr}~input.

%% file: 5-Conclusions/Conclusions.tex
\section{Conclusions}

This paper proposes an extension to the \gls*{mprnet}~\cite{mehri2021mprnet} architecture that achieves better performance on \gls*{lp} super-resolution by combining \gls*{ps} layers and attention modules.
We proposed a new perceptual loss that combines the recognition results achieved by an \gls*{ocr} model with the \gls{ssim} metric between the \gls*{lr} and \gls*{hr} images for better \gls*{lp}~reconstruction.

The main intuition behind our approach is to exploit channel reorganization and learning capabilities from the \gls*{ps} and \gls*{pu} layers for custom scale operations instead of translational invariance and interpolation methods.
An autoencoder with \gls{ps} and \gls{pu} layers for shallow feature extraction was added as an input block to generate an attention mask with regions of interest within the image for reconstruction.
Thus, by aggregating the mask and original input, we optimized computational resources and generated \gls{sr} images with emphasis on relevant information. %
\gls*{pltfam} was proposed to better explore inter-channel relationship features and their position within the image.
We exploited the \gls*{ps} and \gls*{pu} layers instead of the original \gls*{tfam} \textit{MaxPool}, \textit{AvgPool} and upscaling~ones.

All of our experiments were conducted on a public dataset with a wide variety of \gls*{lp} images.
The results showed that recognition rates higher than those achieved by two baselines are achieved in images reconstructed by the proposed method.

In future work, we plan to build a large-scale dataset for \gls*{lp} super-resolution containing thousands of pairs of \gls*{lr} and \gls*{hr} images.
More specifically, we aim to collect videos where the \gls*{lp} is perfectly legible on one frame but illegible on another.
In this way, it would be possible to assess current approaches in real-world scenarios, as well as to develop new~methods.
We also intend to carry out experiments in cross-dataset setups to assess and eliminate the impact of dataset bias~\cite{torralba2011unbiased,laroca2022first}.

%% file: 0-aux/acknowledgments.tex
\section*{\uppercase{Acknowledgments}}

\iffinal

    This work was supported in part by the Coordination for the Improvement of Higher Education Personnel~(CAPES) (\textit{Programa de Coopera\c{c}\~{a}o Acad\^{e}mica em Seguran\c{c}a P\'{u}blica e Ci\^{e}ncias Forenses \#~88881.516265/2020-01}), in part by the National Council for Scientific and Technological Development~(CNPq) (\#~308879/2020-1 and \# 309953/2019-7), and also in part by the Minas Gerais Research Foundation (FAPEMIG) (Grant~PPM-00540-17).
    We gratefully acknowledge the support of NVIDIA Corporation with the donation of the Quadro RTX $8000$ GPU used for this research.

\else
    Acknowledgments will be included in the final~version.
\fi